\documentclass[runningheads]{llncs}
\pdfoutput=1

\usepackage[T1]{fontenc}
\usepackage{xcolor}

\definecolor{route}{HTML}{1B9E77}    
\definecolor{retrieve}{HTML}{D95F02} 
\definecolor{reflect}{HTML}{7570B3}  
\definecolor{repair}{HTML}{E7298A}   
\usepackage{graphicx}
\usepackage{amsmath}   
\usepackage{amssymb}   
\usepackage{amsfonts}  
\usepackage{mathtools} 
\usepackage{hyperref}  
\usepackage{algorithmic}
\usepackage{amsmath,amssymb}

\usepackage{array}      
\usepackage{booktabs}   

\usepackage{multirow}

\usepackage{tikz}
\usepackage{pgfplots}
\usepgfplotslibrary{groupplots}
\pgfplotsset{compat=1.18}

\pgfplotsset{every axis/.append style={unbounded coords=jump}}

\newcommand{\OvrlMedGemmaOne}{5.99}   
\newcommand{\OvrlMedGemmaTwo}{6.75}   
\newcommand{\OvrlMedGemmaThree}{7.20}

\newcommand{\OvrlQwenOne}{6.91}       
\newcommand{\OvrlQwenTwo}{6.86}       
\newcommand{\OvrlQwenThree}{7.44}

\newcommand{\OvrlGeminiOne}{7.09}     
\newcommand{\OvrlGeminiTwo}{7.88}     
\newcommand{\OvrlGeminiThree}{8.02}

\newcommand{\MapMedGemmaOne}{5.97}
\newcommand{\MapMedGemmaTwo}{6.50}
\newcommand{\MapMedGemmaThree}{6.75}

\newcommand{\MapQwenOne}{7.65}
\newcommand{\MapQwenTwo}{7.73}
\newcommand{\MapQwenThree}{7.88}

\newcommand{\MapGeminiOne}{10.11}
\newcommand{\MapGeminiTwo}{10.78}
\newcommand{\MapGeminiThree}{10.97}

\usepackage{tikz}
\usepackage{eso-pic}

\begin{document}
\title{%
  \textcolor{route}{Route}, \textcolor{retrieve}{Retrieve}, \textcolor{reflect}{Reflect}, \textcolor{repair}{Repair}: Self-Improving Agentic Framework for Visual Detection and Linguistic Reasoning in Medical Imaging}
%
%
\author{Md. Faiyaz Abdullah Sayeedi\inst{1}\orcidID{0009-0007-3079-7806} \and
Rashedur Rahman\inst{1}\orcidID{0000-0003-0267-2612} \and
Siam Tahsin Bhuiyan\inst{1}\orcidID{0009-0000-1298-1991} \and Sefatul Wasi\inst{1}\orcidID{0009-0004-6949-0607} \and Ashraful Islam\inst{1}\orcidID{0000-0003-2367-2013} \and Saadia Binte Alam\inst{1}\orcidID{0009-0007-0358-7635} \and AKM Mahbubur Rahman\inst{1}\orcidID{0000-0001-9941-4817}}
\authorrunning{M.F.A. Sayeedi et al.}
%
\institute{\textsuperscript{1} Center for Computational \& Data Sciences (CCDS), \\
Independent University, Bangladesh (IUB) \\
\email{msayeedi212049@bscse.uiu.ac.bd, rashed@iub.edu.bd, siamtbhuiyan@gmail.com, sefatulwasi@gmail.com, ashraful@iub.edu.bd, saadiabintealam@gmail.com, akmmrahman@iub.edu.bd}}
\maketitle              

\begin{abstract}
Medical image analysis increasingly relies on large vision--language models (VLMs), yet most systems remain single-pass black boxes that offer limited control over reasoning, safety, and spatial grounding. We propose \emph{R$^4$}, an agentic framework that decomposes medical imaging workflows into four coordinated agents: a \textcolor{route}{Router} that configures task- and specialization-aware prompts from the image, patient history, and metadata; a \textcolor{retrieve}{Retriever} that uses exemplar memory and pass@${k}$ sampling to jointly generate free-text reports and bounding boxes; a \textcolor{reflect}{Reflector} that critiques each draft--box pair for key clinical error modes (negation, laterality, unsupported claims, contradictions, missing findings, and localization errors); and a \textcolor{repair}{Repairer} that iteratively revises both narrative and spatial outputs under targeted constraints while curating high-quality exemplars for future cases. Instantiated on chest X-ray analysis with multiple modern VLM backbones and evaluated on report generation and weakly supervised detection, R$^4$ consistently boosts LLM-as-a-Judge scores by roughly $+1.7$--$+2.5$ points and mAP$_{50}$ by $+2.5$--$+3.5$ absolute points over strong single-VLM baselines, without any gradient-based fine-tuning. These results show that agentic routing, reflection, and repair can turn strong but brittle VLMs into more reliable and better grounded tools for clinical image interpretation. Our code can be found at: \url{https://github.com/faiyazabdullah/MultimodalMedAgent}
\end{abstract}
\section{Introduction}
Medical image analysis plays a central role in modern clinical workflows, from triaging chest radiographs in the emergency department to monitoring tumor response over time \cite{pinto2023artificial}. Recent advances in large vision--language models (VLMs) have enabled impressive zero-shot and few-shot performance on tasks such as report generation, visual question answering, and abnormality detection \cite{LI2025102995}. However, most existing systems still operate as monolithic black boxes: they take an image (and possibly a short prompt) as input, return a single free-text output, and offer limited control over how reasoning is performed, how errors are detected, or how outputs are grounded in concrete image regions. This lack of structure raises concerns about hallucinations, subtle clinical errors, and poor integration with existing radiology workflows, where localization, explanation, and safety checks are essential \cite{KhaBid_HallucinationAware_MICCAI2025}. At the same time, real-world clinical settings are heterogeneous. Different patient groups (for example, oncology patients versus cardiac patients), imaging modalities, and institutions often require distinct reporting conventions, risk thresholds, and safety constraints \cite{xue2025enhancing}. A single “one-size-fits-all’’ prompt for a large model is unlikely to perform optimally across these contexts. Instead, there is a growing interest in \emph{agentic} \cite{lin2025a} solutions that decompose complex tasks into coordinated steps, where specialized components route cases, retrieve relevant priors, critique intermediate outputs, and iteratively refine them. Such multi-agent approaches have shown promise in other domains, but their application to medical imaging remains under-explored, especially when we consider both global reasoning (free-text reports) and local reasoning (bounding boxes and regions of interest) \cite{berceanova}.

In this work, we propose an agentic framework, denoted R$^4$ (\textcolor{route}{Router}, \textcolor{retrieve}{Retriever}, \textcolor{reflect}{Reflector}, and \textcolor{repair}{Repairer}), for self-improving vision--language reasoning \cite{zhang2025agentic} in medical image analysis. The system takes as input a medical image $x$, an optional textual query $q$, and a structured patient context, including a medical history vector and examination metadata. A \emph{Router} agent first selects a task configuration and LLM specialization tailored to the case, for example, favoring a chest-radiology-focused configuration for chest X-rays or an oncology-focused configuration for follow-up CT scans. Conditioned on this decision and on a memory of past cases, a \emph{Retriever} agent generates multiple candidate drafts of the clinical report or answer and, in parallel, produces bounding boxes that localize suspected abnormalities. To improve reliability and transparency, a dedicated \emph{Reflector} agent then critiques each draft--box pair, focusing on clinically important failure modes such as incorrect negation, laterality swaps, unsupported claims, contradictions, missing key findings, and misaligned localization. These critiques are encoded as a structured issue list that directly feeds a \emph{Repairer} agent, which iteratively updates both the free-text report and the bounding boxes under targeted constraints, performing a small number of reflect--repair steps until no material issues remain or a maximum number of iterations is reached.

Our main contributions are as follows:
\begin{itemize}
    \item We introduce an agentic architecture for medical image analysis that explicitly integrates patient history and examination metadata into a Router, which selects a task configuration and LLM specialization rather than relying on a single, static prompt.
    \item We couple global report generation with quantitative localization by designing a Retriever that jointly produces free-text clinical drafts and bounding boxes, and a Reflector--Repairer loop that critiques and refines both the textual reasoning and the spatial annotations.
    \item We propose a persistent exemplar memory that stores high-quality past cases with task, specialization, cues, and tags, and show how this memory can be used to retrieve context-aware few-shot exemplars, enabling self-improvement over time without retraining the underlying vision--language model.
\end{itemize}

\section{Related Work}

Recent work has begun to explore multi-agent paradigms for medical AI. MedAgents \cite{tang-etal-2024-medagents} introduces a role-playing framework where multiple LLM-based “experts’’ iteratively discuss, critique, and refine zero-shot medical question answering, showing that structured collaboration can unlock latent medical knowledge in general-purpose LLMs across MedQA, MedMCQA, PubMedQA, and MMLU subsets. MMedAgent \cite{li-etal-2024-mmedagent} extends this idea to tool use, training an agent to select among specialized medical tools across several modalities and tasks, and demonstrating that learned tool orchestration can outperform both open-source and proprietary baselines such as GPT-4o. PathFinder \cite{ghezloo2025pathfindermultimodalmultiagentmedical} takes a complementary view in histopathology, using a triage, navigation, description, and diagnosis agent pipeline to emulate how pathologists traverse whole-slide images, achieving both accuracy gains and interpretable multi-scale reasoning.

Closer to radiology, Yi et al.\ propose a multimodal multi-agent framework for radiology report generation that decomposes the workflow into retrieval, draft generation, visual analysis, refinement, and synthesis agents, improving both automatic metrics and LLM-as-a-Judge evaluations \cite{yi2025multimodal}. Medical AI Consensus \cite{boardy2025medical} frames radiology report generation and evaluation as a multi-agent reinforcement learning environment, where specialized agents produce, review, and score reports, and LLMs act as evaluators alongside radiologists. These works show that breaking monolithic report generation into coordinated agent roles can improve structure and clinical plausibility, but they typically focus on text-only evaluation or coarse image-to-text alignment, with less emphasis on explicit spatial grounding via bounding boxes \cite{SUN2025108870}.

A parallel line of work studies how to better align multimodal models with radiology workflows without explicit multi-agent decompositions. Gla-AI4BioMed \cite{zhang-etal-2024-gla} aligns CLIP features with a Vicuna-based LLM through visual instruction tuning, achieving strong performance on chest X-ray report generation and highlighting the importance of faithfully modeling findings and impressions. Rad-Flamingo \cite{anonymous2025radflamingo} proposes a multimodal, prompt-driven framework that integrates images and clinical notes to generate reports with patient-centric explanations, including a synthetic pipeline for augmenting benchmarks with lay summaries. BoxMed-RL \cite{JING2026103910} focuses on verifiable report generation by combining chain-of-thought reasoning with reinforcement learning to encourage spatial grounding, improving both text metrics (METEOR, ROUGE-L) and LLM-based scores on MIMIC-CXR and IU X-Ray, but requires non-trivial training and policy optimization.

Our R$^4$Agent framework is complementary to these efforts. Like MedAgents, it uses an agentic decomposition to orchestrate LLM/VLM capabilities, but it is explicitly designed for \emph{medical image analysis} with joint text--box outputs, specialization-aware routing, and exemplar memory. Compared to prior multi-agent systems \cite{yi2025multimodal,boardy2025medical}, R$^4$ Agent introduces (i) a unified Route $\to$ Retrieve $\to$ Reflect $\to$ Repair loop that operates on both reports and bounding boxes, (ii) a pass@${k}$ draft selection strategy driven by structured, clinically targeted issue lists, and (iii) a lightweight self-improvement mechanism that updates an exemplar memory without any gradient-based fine-tuning. In contrast to training-heavy methods such as BoxMed-RL \cite{JING2026103910} or CLIP+LLM finetuning \cite{zhang-etal-2024-gla}, our approach treats the underlying VLM as a frozen backbone and instead improves reliability, spatial grounding, and clinical fidelity through agentic control and reflective revision.

\section{Methodology}

\subsection{Problem Formulation}

We consider a medical image analysis setting where each sample consists of a medical image $x \in \mathcal{X}$ (for example, a chest radiograph) and an optional textual query $q \in \mathcal{Q}$ (for example, the corresponding clinical reporting). In addition, we assume access to a patient history vector $h_{\mathrm{pat}}$ (such as prior diagnoses, procedures, or medications) and a metadata vector $z$ (such as modality, body region, acquisition site, and basic demographics). The goal is to produce a structured output
\begin{equation}
    y = \bigl(r, B\bigr),
\end{equation}
where $r$ is a free-text clinical report, and $B = \{ b_n \}_{n=1}^{N}$ is a set of bounding boxes that localize clinically relevant regions such as organs and abnormalities. Each bounding box $b_n$ is represented as
\begin{equation}
    b_n = \bigl(\ell_n, d_n, \alpha_n, x^{\min}_n, y^{\min}_n, x^{\max}_n, y^{\max}_n\bigr),
\end{equation}
where $\ell_n$ is a short label (for example, “heart” or “right effusion”), $d_n$ is a brief textual description, $\alpha_n \in [0,1]$ is a confidence score, and $(x^{\min}_n, y^{\min}_n, x^{\max}_n, y^{\max}_n) \in [0,1]^4$ are normalized coordinates in the image reference frame. We implement this mapping using an agentic system composed of four agents: \textcolor{route}{Router}, \textcolor{retrieve}{Retriever}, \textcolor{reflect}{Reflector}, and \textcolor{repair}{Repairer}, collectively denoted as R$^4$. The overall system defines a mapping
\begin{equation}
    \bigl(x, q, h_{\mathrm{pat}}, z, \mathcal{M}\bigr)
    \;\xrightarrow{\;\text{R}^4\;}
    \bigl(\hat{r}, \hat{B}, \mathcal{M}'\bigr),
\end{equation}
where $\mathcal{M}$ is an exemplar memory and $\mathcal{M}'$ is the updated memory after self-improvement.

\begin{figure}[t]
    \centering
    \includegraphics[width=\textwidth]{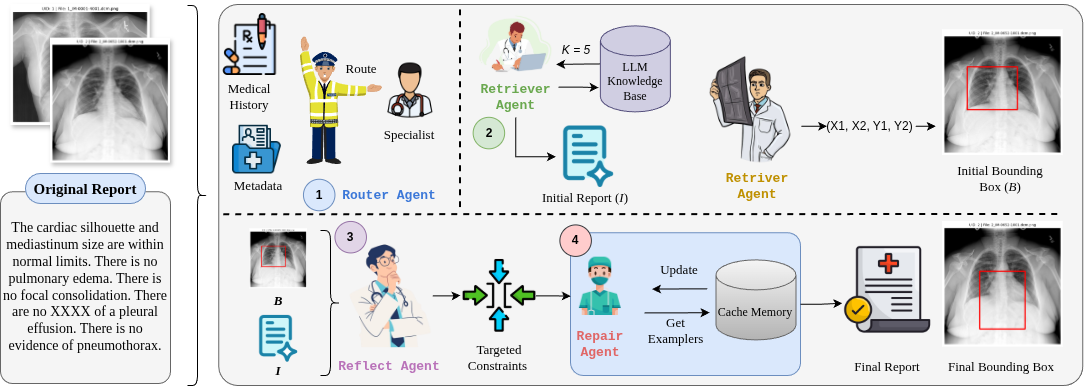}
    \caption{Overview of the proposed R$^4$ agentic framework. The chest X-ray and its accompanying text query are used as input. Patient medical history and metadata are then passed to the Router Agent (1), which selects a task configuration and specialized LLM to handle the case. The Retriever Agent (2) queries the LLM knowledge base with $k$ retrieved exemplars to generate an initial report $I$ together with an initial bounding box $B$ in normalized coordinates $(x_1,x_2,y_1,y_2)$. The Reflect Agent (3) critiques $I$ and $B$ under targeted clinical and safety constraints, producing an issue set $I$. The Repair Agent (4) updates the draft report and bounding box using these constraints, consults the cache memory for similar exemplars, and outputs a final report and refined bounding box that are fed back into the cache for future retrieval.}
    \label{fig:r4_methodology}
\end{figure}

\subsection{R$^4$ Agentic Architecture}

\subsubsection{Router}

Given the input $(x, q)$, we extend the context with patient-level information $h_{\mathrm{pat}}$ and metadata $z$. We first construct a lightweight hint representation
\begin{equation}
    \tilde{h}
    =
    \bigl(q, h_{\mathrm{pat}}, z, \text{task\_hint}, \text{mode\_hint}\bigr),
\end{equation}
where $\text{task\_hint}$ encodes any a priori information about the intended downstream task (report generation versus risk stratification) and $\text{mode\_hint}$ encodes any preferred prompting style (zero-shot, few-shot, chain-of-thought). The Router agent maps this combined hint $\tilde{h}$ to a routing decision
\begin{equation}
    r = \bigl(s, m, F\bigr)
    = \text{Router}(\tilde{h}),
\end{equation}
where $s \in \mathcal{S}$ denotes the selected LLM specialization, $m \in \{zero, few, cot\}$ denotes the prompting mode, and $F$ is a set of flags indicating additional constraints such as whether strict safety checking, bounding-box localization, or longitudinal comparison is required. The specialization index $s$ effectively chooses a tailored configuration for the downstream vision--language model, for example with a chest-radiology focus, oncology follow-up focus, or cardiovascular risk focus, and with prompt templates and tool access adapted to the patient’s history and metadata. In practice, we first compute a heuristic decision $\hat{r} = (\hat{s}, \hat{m}, \hat{F})$ by applying simple rules to $(h_{\mathrm{pat}}, z)$, such as mapping a known oncology history and CT modality to an oncology-specialized configuration, and then optionally refine this decision using a small vision--language model call that returns $r$ in a structured JSON form. If this model call fails or its response cannot be parsed, the heuristic decision $\hat{r}$ is used as a fallback. In this way, the Router serves as a high-level controller that selects which specialized LLM configuration should analyze the data, conditioned explicitly on the patient’s medical history and examination metadata.

\subsubsection{Exemplar Memory}

We maintain a persistent memory $\mathcal{M}$ containing previously processed cases \cite{xu2025amem}, defined as
\begin{equation}
    \mathcal{M} = \bigl\{ m_i \bigr\}_{i=1}^{K},
    \quad
    m_i = \bigl(t_i, s_i, c_i, f_i, \text{tags}_i\bigr),
\end{equation}
where $t_i$ denotes the task type of the $i$-th exemplar (such as chest X-ray reporting or longitudinal follow-up), $s_i \in \mathcal{S}$ denotes the LLM specialization that handled that case, $f_i$ is the final answer or report, $c_i$ is a short textual cue summarizing the case, and $\text{tags}_i$ is a small set of discrete tags encoding salient properties such as the presence of comorbidities, specific findings, or safety-relevant phenomena (for example, negation, laterality, oncology\_history). Given a new case, we derive a cue $c$ by combining information from the query $q$, the patient history vector $h_{\mathrm{pat}}$, and the metadata $z$, or by extracting key entities from an intermediate draft when available. We then compute a lexical overlap score
\begin{equation}
    s(m_i, c) =
    \bigl\lvert
        W(c) \cap \bigl( W(c_i) \cup \text{tags}_i \bigr)
    \bigr\rvert,
\end{equation}
where $W(\cdot)$ denotes the set of lowercased tokens. Restricting attention to exemplars that match the current task $t$ and Router-selected specialization $s$, we select the top-$k$ items
\begin{equation}
    \text{TopK}(c, t, s) =
    \operatorname*{arg\,top}_i^{k} s(m_i, c)
    \quad
    \text{subject to } t_i = t,\, s_i = s,
\end{equation}
and use these retrieved exemplars as few-shot context in the Retriever. In this way, the same R$^4$ framework can adapt its behavior to different clinical phenotypes and workflows by retrieving examples that are simultaneously task-aligned, specialization-aligned, and patient-context–aligned.

\subsubsection{Retriever: Draft Generation and Bounding Boxes}

Conditioned on the routing decision $r$ and the input $(x, q)$, the Retriever agent generates $k$ candidate drafts using a pass@${k}$ strategy. For each pass $j \in \{1,\dots,k\}$, it produces
\begin{equation}
    \bigl(d_j, \eta_j\bigr)
    = \text{Retrieve}\bigl(x, q, r, \mathcal{M}\bigr),
\end{equation}
where $d_j$ is a textual draft report and $\eta_j$ is metadata. The textual draft $d_j$ is generated by a vision--language model that receives the image $x$, a task- and specialization-specific prompt $p(t, s, m)$ that reflects the chosen task and prompting mode, and an optional set of few-shot exemplars obtained from $\text{TopK}(c, t, s)$. In parallel, the Retriever invokes a quantitative localization sub-agent that acts as a bounding-box detector:
\begin{equation}
    B_j = \text{BBoxAgent}\bigl(x, d_j, q\bigr).
\end{equation}
This BBoxAgent is prompted to return a list of bounding boxes, where each element $b_n$ follows the parameterization $b_n = (\ell_n, d_n, \alpha_n, x^{\min}_n, y^{\min}_n, x^{\max}_n, y^{\max}_n)$ with normalized coordinates in $[0,1]$. The metadata $\eta_j$ therefore contains
\begin{equation}
    \eta_j = \bigl(\text{fewshots\_used}_j, m, t, s,
    \text{temperature}_j, B_j\bigr),
\end{equation}
which records the number of exemplars, the prompting mode, the task and specialization, the sampling temperature, and the set of bounding boxes.

\subsubsection{Reflector: Issue Detection}

For each draft $d_j$, the Reflector agent analyzes the pair $(x, d_j)$, together with $q$ if available, and returns a structured list of issues:
\begin{equation}
    I_j = \text{Reflect}(x, d_j, q),
\end{equation}
where
\begin{equation}
    I_j = \bigl\{
        i_{j,1}, i_{j,2}, \dots, i_{j,L_j}
    \bigr\}.
\end{equation}
Each issue $i_{j,\ell}$ includes a type, a location, a message, and a proposed fix. The type field is drawn from a small set that targets clinically important error modes:
\begin{equation}
     \texttt{type} \in
    \{
        \text{negation},
        \text{laterality},
        \text{unsupported},
        \text{contradiction},
        \text{missing}
    \},
\end{equation}
which correspond respectively to failures in negation handling, side (left/right) errors, unsupported claims, internal contradictions, and missing key findings. The Reflector is implemented as a separate vision--language call that is constrained to output a JSON list of such issue records and can be configured via the flags $F$ to emphasize particular safety or coverage constraints.

\subsubsection{Draft Scoring and Selection (pass@k)}

To select the best draft among the $k$ candidates, we define a penalty-based score that aggregates the issues in $I_j$. For each issue $i \in I_j$, we assign a non-negative weight $w_{\texttt{type}(i)}$ that reflects its severity. For example, \texttt{missing} and \texttt{contradiction} errors may receive higher weights than minor unsupported statements. The score for draft $j$ is then
\begin{equation}
    S_j
    =
    - \sum_{i \in I_j} w_{\texttt{type}(i)}.
\end{equation}
A higher score indicates fewer or less severe issues. We select the index
\begin{equation}
    j^\star
    =
    \operatorname*{arg\,max}_{j \in \{1,\dots,k\}} S_j,
\end{equation}
and carry forward the corresponding draft and bounding boxes as
\begin{equation}
    d^{(0)} = d_{j^\star}, \qquad
    \hat{B} = B_{j^\star}.
\end{equation}
Thus, the bounding boxes used in the final output are aligned with the best-scoring textual draft.

\subsubsection{Repairer: Iterative Revision}

Starting from the selected draft $d^{(0)}$ and its associated bounding boxes $B^{(0)} = B_{j^\star}$, the Repairer agent performs a small number of iterative reflect--repair steps on the joint state $(d^{(t)}, B^{(t)})$. At iteration $t$, the Reflector recomputes the issues for the current draft and boxes:
\begin{equation}
    I^{(t)} = \text{Reflect}\bigl(x, d^{(t)}, B^{(t)}, q\bigr),
\end{equation}
where $I^{(t)}$ may include findings related to both textual reasoning (for example, negation or laterality errors) and localization quality (for example, missing or misaligned bounding boxes). If $I^{(t)}$ contains at least one material issue (that is, an issue with type distinct from system or parsing noise), the Repairer produces an updated report and updated boxes
\begin{equation}
    \bigl(d^{(t+1)}, B^{(t+1)}\bigr)
    =
    \text{Repair}\bigl(x, d^{(t)}, B^{(t)}, I^{(t)}, r\bigr),
\end{equation}
where $r$ encodes the routing decision that specifies task $t$, specialization $s$, prompting mode $m$, and flags $F$. This process is repeated for at most $T$ iterations. If at iteration $t$ the Reflector returns no material issues, the loop is stopped early. Let $T^\star \leq T$ denote the last iteration, and define
\begin{equation}
    \hat{r} = d^{(T^\star)}, \qquad
    \hat{B} = B^{(T^\star)}.
\end{equation}
In this formulation, the Repairer jointly refines both the free-text clinical reasoning and the quantitative localization annotations, so that the final report and bounding boxes are consistent with each other and with the underlying image evidence.

\subsection{Self-Improvement via Exemplar Curation}

After obtaining a high-quality final output $(\hat{r}, \hat{B})$ that does not represent a system error, we perform exemplar curation and update the memory $\mathcal{M}$. We define a cue function $g$ that maps the task, query, patient context, and final report to a compact textual cue, and a tag function $\tau$ that extracts simple binary properties from the report:
\begin{align}
    c_{\text{new}} &= g(t, s, q, h_{\mathrm{pat}}, z, \hat{r}), \\
    \text{tags}_{\text{new}} &= \tau(\hat{r}).
\end{align}
For example, $g$ may extract key diagnostic entities such as “cardiomegaly”, “effusion”, or “normal study”, while $\tau$ may add a \texttt{negation} tag if the report contains explicit negated findings, or a \texttt{laterality} tag if left/right terms are present. We then form a new memory item
\begin{equation}
    m_{\text{new}}
    =
    \bigl(t, s, c_{\text{new}}, \hat{r}, \text{tags}_{\text{new}}\bigr),
\end{equation}
and update the memory as
\begin{equation}
    \mathcal{M}' = \mathcal{M} \cup \{ m_{\text{new}} \}.
\end{equation}
In subsequent cases, the updated memory $\mathcal{M}'$ is used within the Retriever to select few-shot exemplars via $\text{TopK}(c, t, s)$, allowing the system to progressively self-improve through better retrieval, without modifying the parameters of the underlying vision--language model.

\section{Experimental Setup}

\subsection{Datasets}

We evaluate the proposed R$^4$ framework on two complementary public chest X-ray datasets. For bounding-box prediction, we use the VinBigData Chest X-ray Abnormalities Detection dataset, which contains 18{,}000 de-identified postero--anterior chest radiographs in DICOM format, each annotated by a panel of radiologists for up to 14 thoracic findings plus a ``No finding'' label.\footnote{\href{https://www.kaggle.com/competitions/vinbigdata-chest-xray-abnormalities-detection}{VinBigData Chest X-ray Abnormalities Detection Dataset}} Each annotation includes a class identifier and a rectangular bounding box, and many images contain multiple objects. For report generation, we use the IU Chest X-rays collection from Indiana University, which provides 7{,}470 chest radiographs and 7{,}414 associated free-text radiology reports with a vocabulary of 1{,}891 unique tokens.\footnote{\url{https://www.kaggle.com/datasets/raddar/chest-xray-images-and-reports} (originally from \url{https://openi.nlm.nih.gov})} The IU images were originally acquired in DICOM format and converted to PNG with basic normalization and resizing; the dataset also distinguishes frontal and lateral projections, and we focus on frontal images for consistency with the VinBigData collection. In all experiments, we prompt models to generate both bounding boxes and reports for images from both datasets, but we evaluate each modality only where ground truth exists: detection is scored on VinBigData, and report generation is scored on IU Chest X-rays.

\subsection{Models}

We consider both closed-source and open-source vision--language models. As a closed-source backbone, we use Gemini-2.5-Flash. As open-source backbones, we include MedGemma-4B, LLaVA-Med, Medical-Llama3-8B, LLaVA-1.5-7B, LLaMA-3.2-11B-Vision, Gemma-3-4B, DeepSeek-VL2, InternVL-4B, VILA-3B, and Qwen2-VL. All backbones are used as frozen black boxes; we do not perform gradient-based fine-tuning. We instantiate two families of systems. The first consists of direct model baselines, where each backbone is prompted once per case in one of three prompting regimes: zero-shot, few-shot, or chain-of-thought (CoT). The second consists of our agentic variants R$^4$Agent-MedGemma-4B, R$^4$Agent-Qwen2.5-VL-7B, and R$^4$Agent-Gemini, which wrap the corresponding backbone inside the full R$^4$ pipeline (Router, Retriever, Reflector, Repairer) with an external memory $\mathcal{M}$ and pass@3 self-improvement.

\subsection{Evaluation Metrics}

For report generation, we evaluate zero-shot and few-shot outputs with BLEU and ROUGE-L, and compute BERTScore (F$_1$) to assess semantic similarity between the generated report $\hat{r}$ and the reference report $r$. To complement these surface- and embedding-based metrics, we further assess the clinical and linguistic quality of the generated reports using an LLM-as-a-judge setup based on GPT-4o-mini. Given the reference report and the model-generated report for the same image, GPT-4o-mini is prompted to rate five key aspects on a 1--10 scale: (i) coverage of key findings, (ii) consistency with the original report, (iii) diagnostic accuracy, (iv) stylistic alignment with typical radiology reporting, and (v) conciseness and clarity. Higher scores indicate better quality. We report per-aspect averages as well as an overall average score across aspects and test cases.

Let $\hat{r}_i$ and $r_i$ denote the generated and reference reports for test case $i$. The judge model takes this pair and, under a fixed evaluation prompt, outputs five scores between 1 and 10:
\begin{equation}
    \bigl(
        s^{\text{cov}}_i,\,
        s^{\text{cons}}_i,\,
        s^{\text{diag}}_i,\,
        s^{\text{style}}_i,\,
        s^{\text{conc}}_i
    \bigr)
    =
    J(\hat{r}_i, r_i),
\end{equation}
where the components correspond respectively to coverage of key findings, consistency with the original report, diagnostic accuracy, stylistic alignment, and conciseness/clarity. We then define a simple overall judge score for case $i$ as
\begin{equation}
    S_i
    =
    \frac{
        s^{\text{cov}}_i
        +
        s^{\text{cons}}_i
        +
        s^{\text{diag}}_i
        +
        s^{\text{style}}_i
        +
        s^{\text{conc}}_i
    }{5},
\end{equation}
and report averages of each component score and of $S_i$ over all test cases. Higher values indicate better judged quality. On the VinBigData Chest X-ray Abnormalities Detection dataset, we compute mean Average Precision (mAP) at IoU 0.5 across the 14 abnormality classes. Given a predicted box $\hat{b}$ and ground-truth box $b$, we use
\begin{equation}
    \text{IoU}(\hat{b}, b)
    =
    \frac{\text{area}(\hat{b} \cap b)}{\text{area}(\hat{b} \cup b)},
\end{equation}
and count a detection as correct if the class label matches and $\text{IoU} \geq 0.5$. For ``No finding'' cases, we additionally track the false-positive rate of spurious abnormal boxes, which directly reflects the model’s ability to remain conservative on normal studies.

\section{Results \& Analysis}

\begin{table}[!ht]
    \centering
    \tiny
    \hspace*{-1cm}
    \begin{tabular}{l|ccc|ccc|ccc|ccccc|c}
    \hline
    \multirow{2}{*}{\textbf{Model / Agent}} &
      \multicolumn{3}{c|}{\textbf{Zero-Shot}} &
      \multicolumn{3}{c|}{\textbf{3-Shot}} &
      \multicolumn{3}{c|}{\textbf{CoT / Agent}} &
      \multicolumn{5}{c|}{\textbf{LLM-as-a-Judge}} &
      \multirow{2}{*}{mAP} \\
    \cline{2-15}
    & BL & BS & RL &
      BL & BS & RL &
      BL & BS & RL &
      Fnd & Sty & Con & Diag & Ovrl & \\
    \hline
    MedGemma-4B
        & 0.0176 & 0.8682 & 0.2036
        & 0.0533 & 0.8780 & 0.2337
        & 0.0408 & 0.8807 & 0.2329
        & 4.49 & 6.55 & 5.48 & 5.40 & 5.48 & 4.285 \\
    LLaVA-Med-7B
        & 0.0201 & 0.8714 & 0.2286
        & \textbf{0.0549} & 0.8812 & 0.2272
        & 0.0476 & 0.8799 & 0.2323
        & 4.54 & 6.67 & 4.98 & 5.62 & 5.45 & 3.852 \\
    Med-Llama3-8B
        & 0.0199 & \textbf{0.8850} & \textbf{0.2379}
        & 0.0520 & \textbf{0.8991} & 0.2378
        & 0.0491 & 0.8836 & \textbf{0.2410}
        & 4.61 & 7.01 & 5.21 & 5.55 & 5.60 & 5.090 \\
    LLaMA-3.2-11B
        & 0.0220 & 0.8656 & 0.2299
        & 0.0491 & 0.8734 & \textbf{0.2425}
        & 0.0407 & 0.8661 & 0.2386
        & 4.33 & 6.93 & 4.70 & 5.04 & 5.25 & 5.630 \\
    Gemma-3-4B
        & 0.0183 & 0.8485 & 0.1994
        & 0.0398 & 0.8577 & 0.2066
        & 0.0420 & 0.8676 & 0.2212
        & 4.17 & 6.32 & 5.29 & 5.35 & 5.28 & 2.729 \\
    Qwen2.5-VL-7B
        & 0.0234 & 0.8251 & 0.2198
        & 0.0466 & 0.8375 & 0.2243
        & 0.0459 & 0.8388 & 0.2367
        & 4.11 & 6.68 & 4.85 & 5.04 & 5.17 & 4.350 \\
    DeepSeek-VL2-3B
        & 0.0175 & 0.7845 & 0.1190
        & 0.0297 & 0.7738 & 0.2378
        & 0.0491 & 0.8836 & 0.2410
        & 4.20 & 5.97 & 5.09 & 4.83 & 5.02 & 1.841 \\
    Gemini-2.5-Flash
        & \textbf{0.0238} & 0.8823 & 0.1593
        & 0.0543 & 0.8758 & 0.2392
        & 0.0511 & 0.8759 & 0.2381
        & 4.58 & 6.73 & 5.54 & 5.47 & 5.58 & 7.490 \\
    \hline
    Z.\ Yi et al.\ \cite{yi2025multimodal}
        & -- & -- & --
        & -- & -- & --
        & 0.0466 & 0.8819 & \textbf{0.2471}
        & 6.36 & 8.16 & 6.74 & 8.26 & 7.38 & -- \\
    \textbf{R$^4$Agent-MedGemma-4B}
        & -- & -- & --
        & -- & -- & --
        & 0.0414 & 0.8767 & 0.1985
        & 6.03 & 7.99 & 6.53 & 8.26 & 7.20 & 6.75 \\
    \textbf{R$^4$Agent-Qwen2.5-VL-7B}
        & -- & -- & --
        & -- & -- & --
        & 0.0437 & 0.8791 & 0.2114
        & 6.44 & 8.14 & 6.86 & 8.31 & 7.44 & 7.88 \\
    \textbf{R$^4$Agent-Gemini}
        & -- & -- & --
        & -- & -- & --
        & \textbf{0.0550} & \textbf{0.8920} & 0.2229
        & \textbf{7.51} & \textbf{8.21} & \textbf{7.27} & \textbf{9.10} & \textbf{8.02} & \textbf{10.97} \\
    \hline
    \end{tabular}
    \caption{Single-VLM and multi-agent results on chest X-ray report generation. BL = BLEU, BS = BERTScore, RL = ROUGE-L. Fnd/Sty/Con/Diag/Ovrl = LLM-as-a-Judge scores (Findings, Style, Consistency, Diagnostic Accuracy, Overall). mAP = mAP$_{50}$ for weakly supervised localization (supervised YOLOv12s reference: \textbf{29.8}). For multi-agent rows (bottom block), Zero-Shot and 3-Shot columns are not applicable (indicated by ``--''), and the CoT/Agent columns report their single report-generation setting. Best value in each column is in bold.}
    \label{tab:merged_en}
\end{table}

Table~\ref{tab:merged_en} reports a unified comparison of single vision--language model (VLM) baselines and multi-agent systems on chest X-ray report generation with weakly supervised localization. For single VLMs, we evaluate three prompting regimes (Zero-Shot, 3-Shot, and CoT). We additionally report mAP$_{50}$ for bounding-box localization of a fully supervised YOLOv12s detector that achieves $29.8$ and serves as an upper-bound reference.

Across the single-VLM baselines, surface-form metrics remain low in absolute BLEU (typically $<0.06$) despite strong semantic similarity, which is expected in radiology due to paraphrasing and variable report structure. Few-shot prompting generally improves token overlap: for example, the best 3-shot BLEU is achieved by LLaVA-Med-7B ($0.0549$), while Medical-Llama3-8B attains the best 3-shot BERTScore ($0.8991$) and the strongest CoT ROUGE-L among single VLMs ($0.2410$). These trends indicate that in-context exemplars can stabilize phrasing and increase overlap, but gains saturate quickly and do not consistently translate into clinically meaningful improvements.

Judge-based evaluation reveals a more clinically relevant picture. Single-VLM overall judge averages cluster narrowly around $\sim 5.0$--$5.6$, with Medical-Llama3-8B achieving the best overall among single-pass models ($5.60$) and Gemini-2.5-Flash close behind ($5.58$). A consistent pattern is that \emph{Style} scores (up to $7.01$) are higher than \emph{Diagnostic Accuracy} (typically $\sim 5.0$--$5.6$), suggesting that many reports are fluent and radiology-like while remaining vulnerable to subtle factual errors, omissions, or contradictions. This mismatch underscores the limitation of relying only on overlap metrics, as clinically unsafe reports can still appear stylistically correct and semantically similar.

Localization is challenging for all single-VLM baselines. The best mAP$_{50}$ among single models is Gemini-2.5-Flash ($7.49$), followed by LLaMA-3.2-11B ($5.63$) and Medical-Llama3-8B ($5.09$), all far below the supervised reference ($29.8$). This gap indicates that single-pass generation struggles to reliably align textual findings (e.g., laterality-specific opacities) with consistent spatial regions, which limits downstream interpretability and trust in clinical workflows.

The multi-agent block demonstrates that agentic decomposition can substantially improve both clinical quality and spatial grounding without changing the underlying VLM parameters. The prior multi-agent baseline of Yi et al.~\cite{yi2025multimodal} achieves a strong CoT/Agent ROUGE-L of $0.2471$ and an overall judge score of $7.38$, already exceeding the best single-VLM judge averages by a wide margin, although it does not report mAP$_{50}$. Our R$^4$Agent variants consistently raise judge scores and improve localization. R$^4$Agent-MedGemma-4B increases the overall judge score from $5.48$ (single MedGemma) to $7.20$ and improves mAP$_{50}$ from $4.285$ to $6.75$. R$^4$Agent-Qwen2.5-VL-7B raises the overall judge score from $5.17$ to $7.44$ and mAP$_{50}$ from $4.35$ to $7.88$, showing that structured reflection and iterative repair help correct clinically critical errors and also refine box-level grounding.

R$^4$Agent-Gemini is the strongest overall configuration. It attains the best CoT/Agent BLEU ($0.0550$) and BERTScore ($0.8920$) across all methods, while also achieving the best judge scores in every sub-dimension (Findings $7.51$, Style $8.21$, Consistency $7.27$, Diagnostic Accuracy $9.10$) and the best overall judge average ($8.02$). Importantly, it also achieves the highest localization score (mAP$_{50}=10.97$), improving over the single Gemini baseline ($7.49$) and the other R$^4$ backbones. While this remains below the supervised detector reference, the consistent jump across backbones indicates that jointly reflecting on and repairing both text and bounding boxes is an effective strategy for improving multimodal grounding in report generation.

Overall, the unified results suggest a stable trend: single-pass VLMs can achieve reasonable semantic similarity but plateau on clinically oriented metrics and struggle with localization, whereas the R$^4$ pipeline systematically boosts clinical reliability (judge averages rising from $\sim 5.2$--$5.6$ to $7.2$--$8.0$) while also improving spatial grounding (mAP$_{50}$ increasing by $+2.5$ to $+3.5$ absolute points across backbones). These gains support the premise that routing, exemplar-conditioned retrieval, structured reflection, and iterative repair provide complementary benefits beyond prompt-only improvements, producing reports that are both more diagnostically accurate and better aligned with localized image evidence.

\begin{figure*}[t]
    \centering
    \begin{tikzpicture}
    \begin{groupplot}[
        group style={group size=2 by 1, horizontal sep=1.2cm},
        width=0.49\linewidth,
        height=0.52\linewidth,
        xmin=1, xmax=3,
        xtick={1,2,3},
        grid=both,
        legend style={at={(0.02,0.98)},anchor=north west,draw=none,fill=none},
        legend cell align=left,
    ]

    \nextgroupplot[
        xlabel={Pass index ($k$ in pass@k)},
        ylabel={LLM-as-a-Judge (Ovrl)},
        ymin=4, ymax=15,
    ]
    \addplot+[mark=*, thick]
        coordinates {(1,\OvrlMedGemmaOne) (2,\OvrlMedGemmaTwo) (3,\OvrlMedGemmaThree)};
    \addlegendentry{R$^4$Agent-MedGemma-4B}

    \addplot+[mark=square*, thick]
        coordinates {(1,\OvrlQwenOne) (2,\OvrlQwenTwo) (3,\OvrlQwenThree)};
    \addlegendentry{R$^4$Agent-Qwen2.5-VL-7B}

    \addplot+[mark=triangle*, thick]
        coordinates {(1,\OvrlGeminiOne) (2,\OvrlGeminiTwo) (3,\OvrlGeminiThree)};
    \addlegendentry{R$^4$Agent-Gemini}

    \nextgroupplot[
        xlabel={Pass index ($k$ in pass@k)},
        ylabel={mAP$_{50}$},
        ymin=0, ymax=15,
    ]
    \addplot+[mark=*, thick]
        coordinates {(1,\MapMedGemmaOne) (2,\MapMedGemmaTwo) (3,\MapMedGemmaThree)};
    \addplot+[mark=square*, thick]
        coordinates {(1,\MapQwenOne) (2,\MapQwenTwo) (3,\MapQwenThree)};
    \addplot+[mark=triangle*, thick]
        coordinates {(1,\MapGeminiOne) (2,\MapGeminiTwo) (3,\MapGeminiThree)};

    \end{groupplot}
    \end{tikzpicture}
    \caption{Pass@k ablation on report quality and localization for three R$^4$Agent backbones. Left: Overall Score LLM-as-a-Judge (Ovrl). Right: localization mAP$_{50}$.}
    \label{fig:passk_ovrl_map}
\end{figure*}
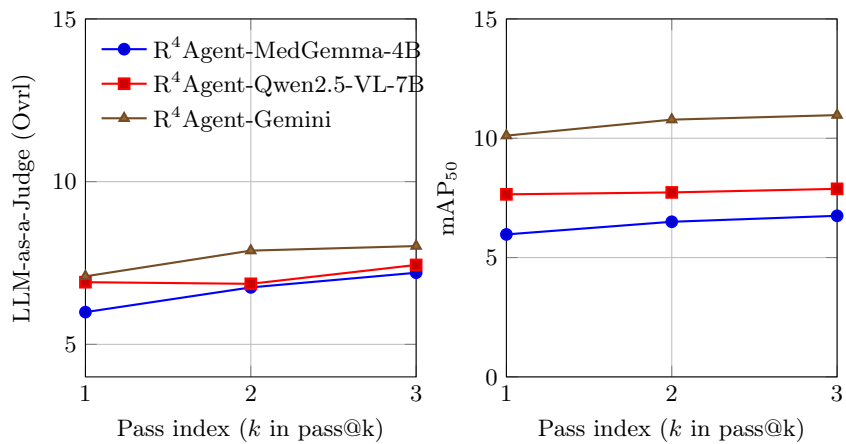

\begin{figure}[t]
    \centering
    \includegraphics[width=\linewidth]{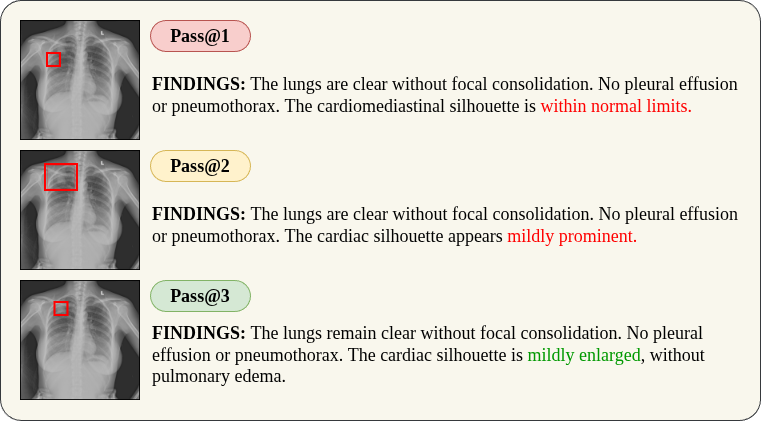}
    \caption{Pass@k qualitative example (backbone: Gemini-2.5-Flash). Report text and bounding-box localization improve from Pass@1 to Pass@3.}
    \label{fig:passk_qualitative}
\end{figure}

\noindent\textbf{Pass@k analysis (Ovrl \& mAP).}
At pass@3, Ovrl reaches $7.20$ (R$^4$Agent-MedGemma-4B), $7.44$ (R$^4$Agent-Qwen2.5-VL-7B), and $8.02$ (R$^4$Agent-Gemini). In parallel, mAP$_{50}$ improves monotonically with additional passes: MedGemma increases from $5.97 \rightarrow 6.50 \rightarrow 6.75$, Qwen from $7.65 \rightarrow 7.73 \rightarrow 7.88$, and Gemini from $10.11 \rightarrow 10.78 \rightarrow 10.97$ (pass@1 $\rightarrow$ pass@2 $\rightarrow$ pass@3). As shown in Fig.~\ref{fig:passk_ovrl_map}, increasing $k$ provides more diverse candidate report--box pairs produced by independent trajectories; pass@k then selects the best candidate using the issue-aware Ovrl scoring. Practically, extra passes help because many radiology failure modes are ``single-shot brittle'': a candidate may miss a key finding or propose an over-broad/misaligned box, and a different pass can avoid that error even with the same backbone and prompts.

The curves in Fig.~\ref{fig:passk_ovrl_map} suggest diminishing returns after $k=2$, especially for localization: the largest jump typically occurs from $k=1$ to $k=2$ (e.g., Gemini $+0.67$ mAP$_{50}$; MedGemma $+0.53$), while $k=2$ to $k=3$ yields smaller but consistent gains (e.g., Gemini $+0.19$; MedGemma $+0.25$). This pattern is expected: early additional passes mainly increase the probability of sampling at least one candidate that is free of major clinical mistakes \emph{and} has a plausible spatial hypothesis, whereas later passes mostly target rarer edge cases and refine residual issues (e.g., tightening a box around the most relevant sub-region). The qualitative example in Fig.~\ref{fig:passk_qualitative} illustrates this behavior: Pass@1 may under-call the cardiac silhouette (e.g., ``within normal limits'') and localize a small, less informative region; Pass@2 moves toward a more appropriate mild prominence with broader coverage; and Pass@3 stabilizes on mild enlargement while maintaining a better-aligned region proposal. Overall, pass@k acts as a lightweight ensembling mechanism over loop outcomes that improves both report correctness (Ovrl) and grounding (mAP$_{50}$) without changing model parameters.

\section{Conclusion}

We introduced R$^4$Agent, an agentic framework that turns general-purpose vision--language models into specialization-aware, self-improving systems for medical image analysis. By combining router-driven configuration, exemplar-guided draft generation, structured clinical reflection, and iterative repair over both text and bounding boxes, R$^4$Agent consistently improves clinical faithfulness and spatial grounding across multiple backbones without any gradient-based fine-tuning. Experiments on chest X-ray report generation and weakly supervised localization show gains of roughly $+1.7$--$+2.5$ points in LLM-as-a-Judge scores and $+2.5$--$+3.5$ absolute mAP$_{50}$ over strong single-VLM baselines, highlighting the value of agentic control beyond raw model scale. In future work, we plan to extend this framework to additional modalities (e.g., CT, MRI, histopathology), incorporate explicit uncertainty calibration, and explore tighter integration with clinician-in-the-loop feedback for deployment in real-world clinical workflows.


%
%
%
\bibliographystyle{splncs04}
\bibliography{mybibliography}

\end{document}